\documentclass[letterpaper, 10 pt, conference]{ieeeconf}  

\IEEEoverridecommandlockouts                              
\usepackage{cite}
\usepackage{amsmath,amssymb,amsfonts}
\usepackage[noend]{algorithmic}
\usepackage{graphicx}
\usepackage{textcomp}
\usepackage{xcolor}
\usepackage{amsmath}

\usepackage[ruled]{algorithm2e}
\usepackage{amsmath} 
\usepackage{booktabs} 
\usepackage{multirow} 
\usepackage{amssymb} 
\usepackage{subfigure}
\usepackage{graphicx}
\usepackage{diagbox}
\usepackage{threeparttable}
 
\usepackage{stackengine}

\usepackage{color}
\usepackage{bm}
\usepackage{hyperref}
\usepackage{marvosym}

\usepackage{colortbl} 
\usepackage{xcolor}
\usepackage{color}

\title{\LARGE \bf
SurgPLAN++: Universal Surgical Phase Localization Network for Online and Offline Inference
}

\author{Zhen Chen$^{1,\dag}$, Xingjian Luo$^{1,\dag}$, Jinlin Wu$^{1,3,*}$, Long Bai$^{2}$, Zhen Lei$^{1,3}$, \textit{Fellow, IEEE},\\Hongliang Ren$^{2}$, \textit{Senior Member, IEEE}, Sebastien Ourselin$^{4}$, Hongbin Liu$^{1,3,*}$, \textit{Senior Member, IEEE}
\thanks{This work is supported by Hong Kong Research Grant Council General Research Fund (GRF) 14211420 and InnoHK program.}%
\thanks{$^{1}$Z. Chen, X. Luo, J. Wu, Z. Lei, H. Liu are with Centre for Artificial Intelligence and Robotics, Hong Kong Institute of Science and Innovation, Chinese Academy of Sciences.}%
\thanks{$^{2}$L. Bai, H. Ren are with The Chinese University of Hong Kong.}%
\thanks{$^{3}$J. Wu, Z. Lei, H. Liu are also with Institute of Automation, Chinese Academy of Sciences.}%
\thanks{$^{4}$S. Ourselin is with King's College London.}%
\thanks{$\dag$\,Equal contribution. $*$\,Corresponding author.}
}

\begin{document}

\maketitle
\thispagestyle{empty}
\pagestyle{empty}


\begin{abstract}
Surgical phase recognition is critical for assisting surgeons in understanding surgical videos. Existing studies focused more on online surgical phase recognition, by leveraging preceding frames to predict the current frame. Despite great progress, they formulated the task as a series of frame-wise classification, which resulted in a lack of global context of the entire procedure and incoherent predictions. Moreover, besides online analysis, accurate offline surgical phase recognition is also in significant clinical need for retrospective analysis, and existing online algorithms do not fully analyze the entire video, thereby limiting accuracy in offline analysis. To overcome these challenges and enhance both online and offline inference capabilities, we propose a universal \underline{Surg}ical \underline{P}hase \underline{L}ocaliz\underline{A}tion \underline{N}etwork, named SurgPLAN++, with the principle of temporal detection. To ensure a global understanding of the surgical procedure, we devise a phase localization strategy for SurgPLAN++ to predict phase segments across the entire video through phase proposals. For online analysis, to generate high-quality phase proposals, SurgPLAN++ incorporates a data augmentation strategy to extend the streaming video into a pseudo-complete video through mirroring, center-duplication, and down-sampling. For offline analysis, SurgPLAN++ capitalizes on its global phase prediction framework to continuously refine preceding predictions during each online inference step, thereby significantly improving the accuracy of phase recognition. We perform extensive experiments to validate the effectiveness, and our SurgPLAN++ achieves remarkable performance in both online and offline modes, which outperforms state-of-the-art methods. The source code is available at \href{https://github.com/franciszchen/SurgPLAN-Plus}{https://github.com/franciszchen/SurgPLAN-Plus}.
\end{abstract}

\section{Introduction}\label{sec1}
The computer-assisted diagnosis and surgery can improve the quality of intervention and facilitate patient healthcare \cite{chen2021super,chen2021diagnose,yang2023hierarchical,chen2022instance,maier2022surgical,xu2024transforming}. In particular, surgical scene understanding \cite{garrow2021machine,zhai2024artificial,chen2024asi} is significant for developing systems to monitor surgical procedures \cite{panesar2020promises}, schedule surgeons \cite{wu2024surgbox}, promote surgical team coordination \cite{chen2024surgfc}, and educate junior surgeons \cite{kirubarajan2022artificial}. 

Surgical phase recognition of surgical videos is challenging and has received great research attention and progress \cite{tecno,ms-tcn,deepphase,trans-svnet}. These studies predominantly focus on online surgical phase recognition to predict the current frame of video streaming without using future frames. Due to the computational burden, these works sequentially extracted spatial and temporal features of surgical videos to advance surgical phase recognition. In this context, most works adopt 2D convolutional neural networks (CNN) to parse each frame, and then adopt diverse temporal mechanisms to exploit the inherent temporal dynamics of surgical videos, \textit{e.g.}, temporal convolution \cite{tecno,ms-tcn}, long short-term memory (LSTM) \cite{deepphase} and transformer \cite{trans-svnet}, generating the phase prediction for the current frame.

\begin{figure*}[t]
  \centering
  \centerline{\includegraphics[width=0.90\textwidth]{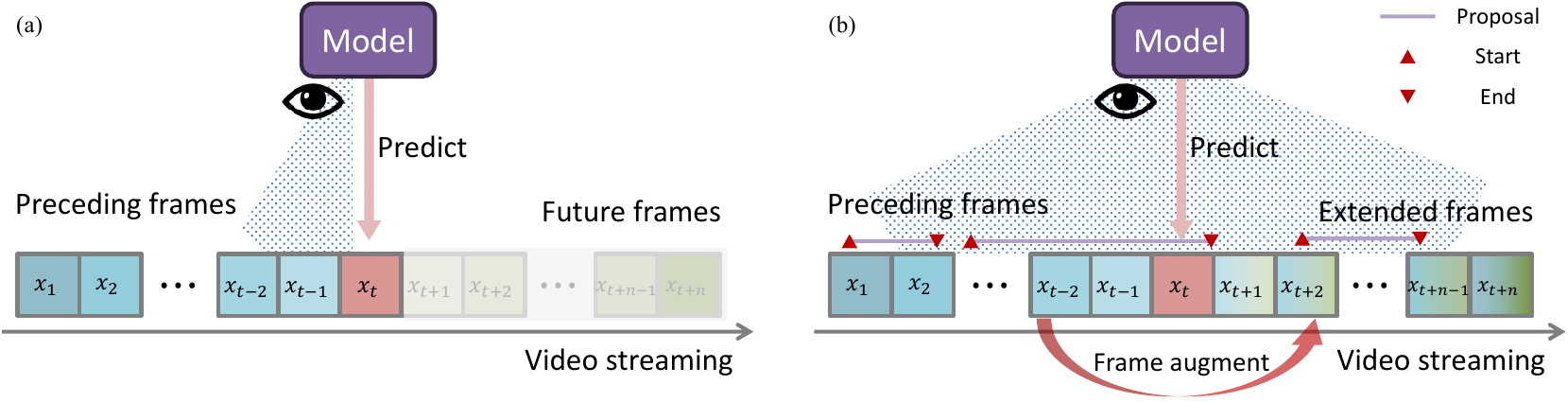}}
\caption{(a) Existing approaches predict as
frame-wise classification, leveraging a certain range of preceding frames. (b) Our SurgPLAN++ extends a pseudo-complete video to generate phase proposals from a global perspective and selects high-quality segments as the surgical phase predictions.}
\label{strategy}
\end{figure*}

Despite great progress in surgical phase recognition, existing works \cite{tecno,ms-tcn,deepphase,trans-svnet} still suffer from two major limitations, including the reliance on frame-by-frame classification and the focus on online analysis to the detriment of offline accuracy. First, existing works formulated the task as a series of frame-by-frame classifications and predicted the current frame by leveraging temporal knowledge from preceding frames. This paradigm,
akin to a greedy strategy, degrades the task of video analysis to a frame-by-frame image prediction task. As illustrated in Fig.~\ref{strategy} (a), these algorithms are unable to conduct global analysis from the perspective of the entire video, resulting in inconsistent predictions of successive frames. Second, these studies merely considered the online analysis of surgical video streaming. In fact, accurate offline surgical phase recognition is also highly desirable with significant clinical needs for retrospective analysis. As a result, these online algorithms are not designed to fully analyze the entire video and could only regard frame-by-frame predictions as the offline surgical phases, thereby leading to inferior accuracy in the offline analysis scenario. In this way, a universal surgical phase recognition framework is highly demanded to analyze the surgical video with a global perspective and is capable of handling both online and offline analysis effectively.

To address these two problems in surgical phase recognition, we propose a universal \underline{Surg}ical \underline{P}hase \underline{L}ocaliz\underline{A}tion \underline{N}etwork, named SurgPLAN++, to enhance both online and offline inference capabilities. As depicted in Fig.~\ref{strategy} (b), the SurgPLAN++ is designed with the principle of temporal detection to ensure a global understanding of the surgical procedure. Specifically, our phase localization strategy first generates phase proposals as starting and ending points from the extracted frame features and then identifies surgical phase segments by filtering the high-confidence proposals. For online analysis, to generate high-quality phase proposals, we devise a data augmentation strategy to extend the streaming video into a pseudo-complete video through diverse augmentations, including mirroring, center-duplication, and down-sampling. For offline analysis, SurgPLAN++ capitalizes on its global phase prediction framework to continuously refine preceding predictions during each online inference step, thereby significantly improving the accuracy of phase recognition. We perform extensive experiments on the Cataract and Cholec80 datasets to validate the effectiveness, and our SurgPLAN++ achieves remarkable performance in both online and offline modes, which outperforms state-of-the-art by a large margin.

\section{Related Work}
\subsection{Video Features Extraction}

The extraction of spatiotemporal features is crucial for video recognition \cite{yang2022semi,chen2023surgical}. Early works used the 3D CNNs to jointly capture spatiotemporal features, such as C3D \cite{C3D}, P3D \cite{P3D}, and I3D \cite{I3D}. However, these approaches encountered a significant challenge: the optimization of temporal and spatial dimensions often conflicted, leading to suboptimal performance. To address this issue, subsequent research proposed a novel divide-and-conquer architecture, SlowFast \cite{slowfast}. This approach employed a dual-branch structure: a low-frame-rate branch for capturing spatial information and a high-frame-rate branch for processing temporal information. SlowFast \cite{slowfast} extracts spatiotemporal information simultaneously, avoiding the optimization conflict between temporal and spatial dimensions, thus achieving improved performance in video recognition.

\subsection{Surgical Phase Recognition}
Surgical phase recognition garnered significant attention in recent years due to its potential to enhance patient safety and streamline surgical workflows. Researchers explored various approaches to automatically identify different phases of surgical procedures. Deep learning models showed promising results in this domain. For instance, PhaseNet \cite{phasenet}, MSTCN \cite{MSTCN}, and TeCNO \cite{tecno} were proposed for recognizing surgical phases by using 2D CNNs. Other studies, such as TMR \cite{TMR}, SV-RCNet \cite{svrcnet} used LSTM to capture temporal dependencies in surgical workflows. Additionally, Transformer-based approaches were explored to improve recognition accuracy, such as Trans-SVNet \cite{trans-svnet}. Despite these advancements, these methods still face systematical challenges in frame-to-frame classification prediction tasks and do not fully leverage the global information provided by the surgical video.


\section{Universal Surgical Phase Localization Network}
\subsection{Overview of SurgPLAN++}\label{overview}
\begin{figure*}[t]
  \centering
  \centerline{\includegraphics[width=1.05\textwidth]{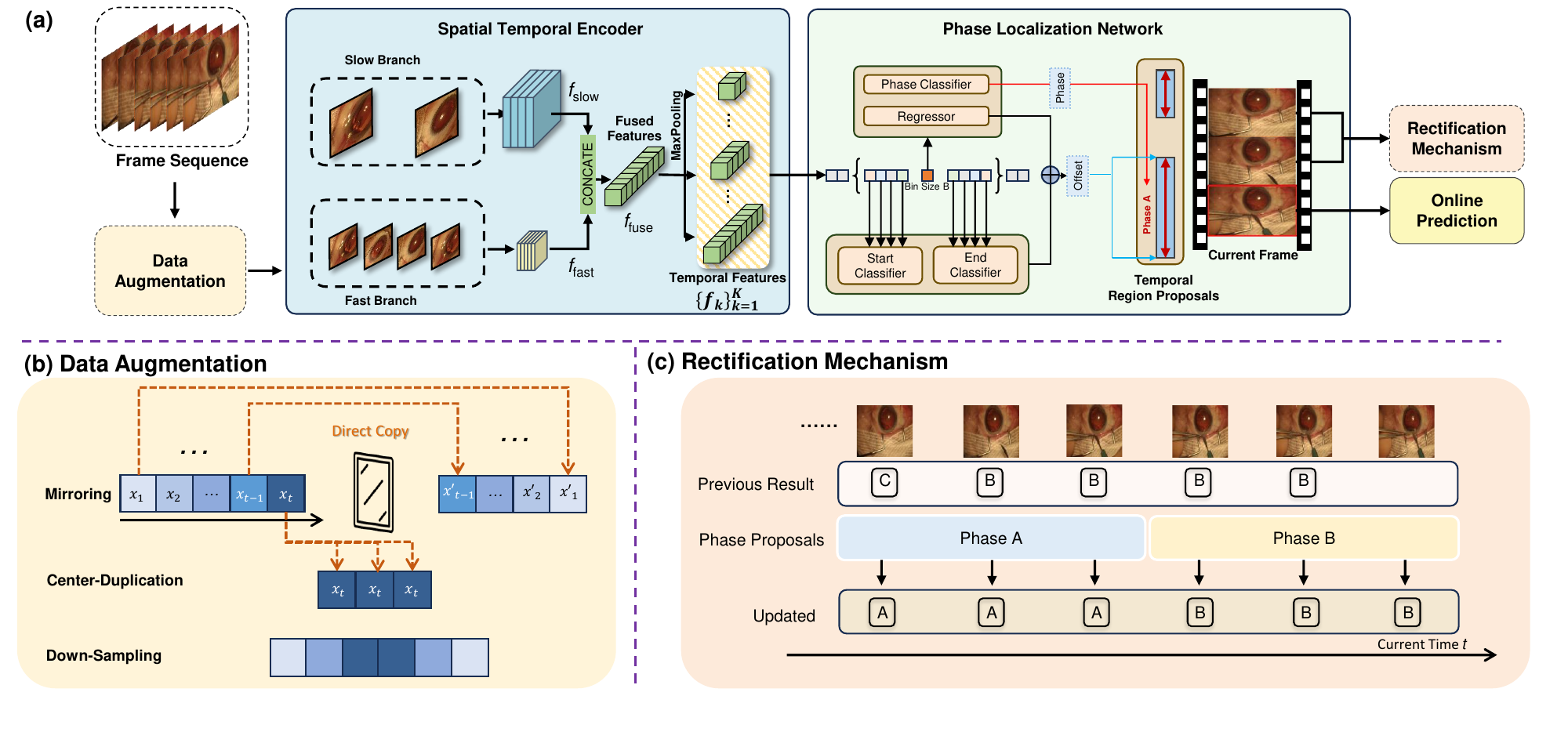}}
\caption{(a) The SurgPLAN++ framework for surgical phase recognition consists of the spatial temporal encoder and phase localization network. (b) In the online mode, the data augmentation extends the streaming video into a pseudo-complete video through mirroring, center-duplication, and down-sampling. (c) In the offline mode, the rectification mechanism further continuously refines preceding predictions during online inference. }
\label{framework}
\end{figure*}

To achieve universal online and offline surgical phase recognition, our SurgPLAN++ is proposed with the temporal detection principle, which consists of a spatial temporal encoder and a phase localization network. As illustrated in Fig.~\ref{framework}, the spatial temporal encoder first extracts multi-scale features of each frame, and then the phase localization network generates phase proposals from frame features and predicts the phase segments as the prediction. 

For the online analysis, SurgPLAN++ utilizes several data augmentations including mirroring, center-duplication, and down-sampling that extend the ongoing video into a pseudo-complete video. For the offline analysis, SurgPLAN++ maintains a dynamic result sequence of phase predictions and updates continuously in each inference step based on the newly proposed segments.

\subsection{Network Architecture}

\noindent \textbf{Spatial Temporal Encoder.} 
We adopt the spatial temporal encoder \cite{slowfast} for SurgPLAN++. The encoder $\bm E$ consists of a slow path and a fast path. The slow path is characterized by a large temporal stride $\mathcal{S}_s$, facilitating the focus on static spatial positional information. Meanwhile, the fast path possesses a small stride $\mathcal{S}_f$, directing attention toward dynamic motion information. Given the surgical video $V\in \mathbb{R}^{T \times H\times W \times 3}$, we generate the slow path features $ f_{\rm slow} \in \mathbb{R}^{\frac{T}{\mathcal{S}_s} \times C_s }$ and fast path features $ f_{\rm fast} \in \mathbb{R}^{\frac{T}{\mathcal{S}_f} \times C_f }$ from two distinct 3D temporal convolutional networks $F_{\rm slow}$ and $F_{\rm fast}$, where $C_s$ and $C_f$ refer to the output feature dimension of the slow and fast path as follows:  
\begin{equation}
\begin{split}
    f_{\rm slow} = F_{\rm slow} (V,\mathcal{S}_s),\\ f_{\rm fast} = F_{\rm fast} (V,\mathcal{S}_f).
\end{split}
\label{slow&fast}
\end{equation}

Then, to concatenate these two features, we utilize a 3D temporal convolution kernel $\mathcal{K}$ to align $f_{\rm fast}$ to the same temporal feature length $\frac{T}{\mathcal{S}_s}$ of the slow path \cite{slowfast}.
\begin{equation}\label{fuse}
    f_{\rm fuse} = [\mathcal{K}(f_{\rm fast}), f_{\rm slow}],
\end{equation}
where $f_{\rm fuse} \in \mathbb{R}^{\frac{T}{\mathcal{S}_s} \times (C_f+C_s) }$ is the fused spatial temporal feature from both paths.

After that, we apply the max-pooling operations with different temporal window sizes $\{w_k\}^K_{k=1}$ to generate spatial temporal features 
$\{f_k\}^K_{k=1}$ at $K$ different scales, as follows:
\begin{equation}\label{maxpooling}
    f_k = \text{MaxPooling}(f_{\rm fuse},w_k), 
\end{equation}
where $f_k$ is the feature sequence processed by a max-pooling layer with the window size $w_k$.  

To this end, the processed features $\{f_k\}^K_{k=1}$ enable the SurgPLAN++ to generate the phase proposals across various scales, thereby enhancing the capability to accommodate differing temporal lengths and improve prediction performance.

\noindent \textbf{Local Start-End Probability.}
The Phase Localization Network $\bm P$ \cite{actionformer,tridet} generates phase proposals that contain the starting and ending points of predicted phases.
The formation of phase proposals is jointly determined by local and global aspects.

First, in local aspect, assume $f^{i}_k$ is temporally the $i^{\rm th}$ feature of the sequence $f_k$, 
we regard this feature as the center of a feature set $\mathcal{F}_k = \{f_k^t\}_{t=i-\frac{B}{2}}^{i+\frac{B}{2}}$ with a predefined bin size $B$. 
We use a regression network $F_{reg}$ to generate conditional start-end distributions $P_s$ and $P_e$ as follows:
\begin{equation}
    P_s,P_e = F_{reg}({f^{i}_k,\mathcal{F}_k,B}).
\end{equation}
Given center point index $i$, the $P_s$ serves as the probability distribution of being the starting point on the left side of the target feature. $P_e$ is the probability distribution for the ending point on the right side of the target feature. The distribution can be expressed as:
\begin{equation}
    \begin{split}
    P_s(l_s|f^{i}_k,B) = \{\ {l_s^t} \mid t\in \{i-\frac{B}{2},i-1\} \  \}, \\
    P_e(l_e|f^{i}_k,B) = \{\ {l_e^t} \mid t\in \{i+1,i+\frac{B}{2}\} \  \},
    \end{split}
\end{equation} 
where $l_s^t$ is the local probability of index $t$ being the starting point and $l_e^t$ is the probability being the ending point.

\noindent \textbf{Global Start-End Probability.} At the global level, the whole feature sequence $f_k$ is processed through three convolution network backbones with similar encoding structures but different linear layer output heads $F_{\rm start}$, $F_{\rm end}$, and $F_{\rm cls}$ to calculate the probabilities of being the starting and ending points $g^i_{s}$ and $g^i_{e}$, and the probability of each phase $g^i_{cls}$, respectively.

\noindent \textbf{Generate Phase Proposals.} 
The predicted start-end point to each target feature $f^i_k$ is calculated by adding the local probability $l^t$ and global probability $g^t$ together as follows:
\begin{equation} \label{distance}
\begin{split}
    \hat{t}_{\rm start}&={\underset{t}{
    \arg\max}}\{l_s^t+g_{\rm s}^{t}\},t\in [i-\frac{B}{2}:i-1], \\
    \hat{t}_{\rm end}&= {\underset{t}{
    \arg\max}}\{l_e^t+g_{\rm e}^{t}\}, t \in [i+1:i+\frac{B}{2}],\\
\end{split} 
\end{equation}
where $\hat{t}_{\rm start}$ and $\hat{t}_{\rm end}$ are  regarded as the boundary of the phase segment.  $[\hat{t}_{\rm start},\hat{t}_{\rm end}]$ is
the proposed phase segment of temporal index $i$ in the feature sequence $f_k$. 

This process is conducted among all the feature sequences $\{f_k\}^K_{k=1}$ generated from the different Max-pooling window sizes to ensure robustness in segment lengths. 
After regional proposals are generated, we apply the Non-Maximum-Suppression (NMS) \cite{softnms} method to filter the generated proposals.
The remaining $M$ proposals $\{Y_m\}^M_{m=1} $ are the predicted phase segments. 

\begin{algorithm}[t]
	\caption{{The training pipeline of SurgPLAN++.}} 
	\SetKwInOut{Input}{\textbf{Input}}\SetKwInOut{Output}{\textbf{Output}} 
	\hspace*{0.02in}\Input{Video $\bm V$ and its phase annotation $\bm T$;\\
	    \hspace*{0.02in}The Spatial Temporal Encoder $\bm{E}$;\\
		\hspace*{0.02in}The Phase Localization Network $\bm{P}$;\\
	}
	\hspace*{0.02in}\Output{The trained Phase Localization Network $\bm{P}$.
	}
	
    \vspace{2.0pt}
	\hspace*{0.02in}{\bf Training:}
	\begin{algorithmic}[1]
		\STATE Load pre-trained Spatial Temporal Encoder $\bm{E_v}$ and initialize the Phase Localization Network $\bm{P}$;\\
		\STATE Generate visual feature $\{f_k\}^K_{k=1}$ from the video $\bm V$ through $\bm{E_v}$ in Eq.\,\eqref{slow&fast}, Eq.\,\eqref{fuse} and Eq.\,\eqref{maxpooling};\\
        \FOR{{\rm each scale $k$ in} $K$}
            
        \FOR{{feature set $\mathcal{F}$} in $f_k$}
            \STATE Predict the phase probability;\\
            \STATE Predict the start-end probability  in Eq.\,\eqref{distance};\\ 
            \STATE Generate phase proposals;       
        \ENDFOR
            
        \ENDFOR
            \STATE  Use NMS to filter the phase proposals;
            \STATE Calculate the Cross-Entropy loss for phases and IoU loss for bounding boxes;
        \STATE Optimize the model $\bm P$ through backward propagation. 
	\end{algorithmic} 
\label{training}
\end{algorithm}

\subsection{Online and Offline Phase Prediction}
As the Phase Localization Network requires complete phase segments to effectively generate phase proposals, SurgPLAN++ has data augmentation techniques including mirroring, center-duplication, and down-sampling that extend the ongoing video to a pseudo-complete video. Meanwhile, SurgPLAN++ can effectively take advantage of the global context information to revise past predictions based on its rectification mechanism.

\noindent \textbf{Online Prediction with Data Augmentation.} 
In the surgical video, a symmetrical attribute typically exists in the initial and terminal phases of the video. 
For instance, the entry of surgical instruments serves as the commencement, while their withdrawal signifies the end. 
Therefore, we utilize mirroring to reverse the video, allowing the originally incomplete video to be supplemented with segments generated through mirroring, resulting in a complete video that includes distinct features of both the initial and final stages.

Specifically, for a video stream $V$ that represents a continuous frames set $\{x_1,x_2,\ldots,x_t\}$, where $x_n \in  \mathbb{R}^{H\times W \times 3} $ refers to the frame at specific time $n$ in the frame sequence $V$, $t$ is the current time point. Mirroring the video stream $V$ means that the processed time frame becomes $\{x_1,\ldots,x_{t-1},x_t, x'_{t-1},\ldots,x'_{1}\}$ where $x'_{t}$ is identical to the $x_{t}$. Therefore, we procure a mirrored video sequence $V_p$ centered upon the current temporal juncture, wherein the latter half $V_e = \{x'_{t-1},x'_{t-2},\ldots,x'_{1}\}$ constitutes a retrograde motion of the first half segment $V_s = \{x_1,x_2,\ldots,x_{t-1}\}$.

Additionally, if a given surgical phase at the current moment is incomplete and excessively brief, there is a potential for the phase localization network to overlook this phase, leading to imprecise predictions. To mitigate this issue, we utilize a center-duplicating method to duplicate the current moment, thereby ensuring it attains enough attention for the phase localization network.
We prolong the duration by duplicating the current video frame $x_t$ and inserting them in the middle of the mirrored time frame, which concurrently preserves the action characteristics more effectively. The time frame becomes $V_d = \{V_s,x_t, \ldots,x_t, V_e\}$.

At last, we employ a down-sampling approach when mirroring and center-duplication results in an excessively extended action length. Specifically, slices with a step size $n$ are selected to constrain the action duration. Thus, we get the processed time frame $V_p$.

By applying these three methodologies, we standardize the action lengths within a specified range, optimizing the model’s detection framework and enhancing its detection capabilities. The prediction for the current frame is the phase of the bounding box that includes the center point. 

\begin{algorithm}[t]
	\caption{{The online inference of SurgPLAN++.}} 
	\SetKwInOut{Input}{\textbf{Input}}\SetKwInOut{Output}{\textbf{Output}} 
	\hspace*{0.02in}\Input{Video Stream $\bm V$;\\
	    \hspace*{0.02in}The Spatial Temporal Encoder $\bm{E}$;\\
		\hspace*{0.02in}The Phase Localization Network $\bm{P}$;\\
	}
	\hspace*{0.02in}\Output{Online phase prediction set $\bm {P_r}$ of the video stream $\bm V$.
 }


	\hspace*{0.02in}{\bf Online Inference:}
	\begin{algorithmic}[1]
     \vspace{-1em}
        \FOR{each time step}
	    \STATE Perform mirroring for Video $\bm V$ and duplicating for center point to get augmented Video $V_d$;\\
		\STATE 
        Down-sample $V_d$ into $V_p$;
        \STATE Generating phase proposals $\{Y_m\}^M_{m=1}$ through $\bm P$;
		\IF{the center point of $V_p$ is included in one of the proposals $\{Y_m\}^M_{m=1} $ with phase $y$}
        \STATE The prediction phase $\bm p$ as $y$;\\
        \ELSE 
        \STATE The prediction $\bm p$ as None;\\
        \ENDIF
        \STATE Append $\bm p$ to Online Prediction Set $\bm {P_r}$;
        \ENDFOR
        \RETURN Online Prediction Set $\bm {P_r}$.
	\end{algorithmic}\label{online inference}
\end{algorithm}

 


\noindent \textbf{Offline Prediction with Rectification Mechanism.}
In the context of retrospective amendments to the prior phases, due to our persistent maintenance of a dynamic phase prediction sequence $\rm R_{phase} = \{ y_1,y_2,\dots,y_t \}$, where $y_n$ is the prediction on time $n$. 
We can revise the historical results by leveraging the filtered phase proposals before the current time $t$.  Phase proposals that include or exceed time $t$ are proposals generated related to the augmented data. Therefore, for those completed phase proposals before time $t$, noted as $\{Y'_m\}$, we regard those proposals as already gathering enough information to determine phases, we update those phases by replacing the $y_n$ to the phase of those filtered completed segments that contain time $n$ by $ y'_n \xrightarrow{} y_n $,
where $y'_i$ is the phase prediction in set $\{Y'_m\}$ at the inference step of current moment $t$. 
 By fully utilizing global temporal knowledge, we update the result sequence $\rm R_{phase}$ at each time step to form a better offline performance.

\begin{algorithm}[htbp!]
	\caption{{The offline inference of SurgPLAN++.}} 
	\SetKwInOut{Input}{\textbf{Input}}\SetKwInOut{Output}{\textbf{Output}} 
	\hspace*{0.02in}\Input{Video Stream $\bm V$;\\
	    \hspace*{0.02in}The Spatial Temporal Encoder $\bm{E}$;\\
		\hspace*{0.02in}The Phase Localization Network $\bm{P}$;\\
	}
	\hspace*{0.02in}\Output{Offline phase prediction set $\bm {P_r}$ of the video stream $\bm V$.}
	
	\hspace*{0.02in}{\bf Offline Inference}{\bf :}
	\begin{algorithmic}[1]
      \vspace{-1em}
        \FOR{each time step}
        \STATE Maintain a dynamic phase prediction sequence $\rm R_{phase}$ until current moment $t$;
	    \STATE Perform mirroring for Video $\bm V$ and duplicating for center point to get augmented Video $V_d$;\\
		\STATE Down-sample $V_d$ into $V_p$;
        \STATE Generate phase proposals $\{Y_m\}^M_{m=1}$ through $\bm P$;
        \STATE Filter region proposals that only contains frames before time $t$ as $\{Y'_m\}$;
        \STATE Update the dynamic phase prediction sequence $\rm R_{phase}$ by the current prediction $\{Y'_m\}$;
        \ENDFOR
        \RETURN $\rm R_{phase}$ as Offline Prediction Set $\bm {P_r}$.
	\end{algorithmic}\label{offline inference}
\end{algorithm}

\subsection{Optimzation and Inference}
We summarize the training process of our SurgPLAN++ framework in Algorithm~\ref{training}. We utilize a combination of distinct cross-entropy loss functions and Intersection over Union (IoU) loss to enable the model's multiple heads to perform both temporal proposal bounding box prediction and phase prediction. This dual-task approach facilitates the concurrent optimization of temporal localization and phase classification within the framework of our proposed model architecture. Furthermore, we summarize two different inference modes of our SurgPLAN++ framework in Algorithm~\ref{online inference} and~\ref{offline inference}. This approach enables seamless utilization of different modes under varying circumstances, as both modes employ the same model and undergo identical training processes.

\section{Experiment}
\subsection{Dataset and Implementation Details}
\noindent\textbf{Cholec80 Dataset}. We perform comparisons on the Cholec80 dataset \cite{twinanda2016endonet} of laparoscopic cholecystectomy procedures, which is the mainstream benchmark for surgical phase recognition. The Cholec80 dataset contains 80 surgical videos with a resolution of $854\times 480$ or $1, 920\times 1, 080$ at 25 frame-per-second (FPS). The laparoscopic cholecystectomy procedures are divided into seven surgical phases. We exactly follow the standard splits \cite{twinanda2016endonet,trans-svnet}, \textit{i.e.}, the first 40 videos for training and the rest 40 videos for test. 

\noindent \textbf{Cataract Dataset}. We further conduct our experiment on the public Cataracts \cite{dataset} dataset and follow the standard split \cite{al2019cataracts} to divide 25 cataract surgery videos for training and the remaining 25 videos for test. These cataract surgery videos are captured with the resolution of $1,920\times1,080$ at 30 FPS. The Cataracts dataset contains 19 phase categories, including one background category without clear surgical purposes. 

\noindent \textbf{Implementation Details}.
We perform the experiments using PyTorch on a single NVIDIA A800 GPU. All videos are resized to $256\times256$ with 1 FPS after preprocessing.
In the training phase of the Phase Localization Network, the learning rate is configured to 0.001, and the Adam optimizer is utilized for the optimization process. 
For our SurgPLAN++ framework, we transform frame-by-frame labels into segments of surgical phases, and each segment consists of the start time, end time, and phase label. The window sizes for the Max-Pooling of fused features are 1, 2, and 4. The bin size is set as 24 in the Cataract \cite{al2019cataracts} dataset. These parameters are chosen because of the statistical information \cite{luo2024surgplan} we collect from the dataset. Since most of the phase lengths are around 0 to 40 seconds, along with the Max-Pooling window size, the bin can cover one complete phase in almost any circumstance.
In the inference stage, the threshold is set to 0.15. Concurrently, for data augmentation in the online mode, the number of feature replications is established at 16 which can make the phase length closer to the real phase length in most of the cases. The scaling ratio will be adjusted to ensure the video length will not exceed 512.

\noindent \textbf{Evaluation Metrics}. We adopt four commonly used metrics to comprehensively evaluate the performance of surgical phase recognition, including accuracy (AC), precision (PR), recall (RE), and Jaccard (JA). Higher scores for these metrics indicate better quality of surgical phase recognition. Following the evaluation protocol in previous works \cite{trans-svnet}, we evaluate the selected state-of-the-art methods under the same criteria as the SurgPLAN++ to perform fair comparisons.

\begin{table}[t]
    \begin{center}
            \caption{Comparison with with state-of-the-arts Cholec80 datasets}
        \begin{tabular}{l| p{1cm}<{\centering} p{1cm}<{\centering} p{1cm}<{\centering} p{1cm}<{\centering} } \hline 
        \toprule[1pt]
            \multirow{1}{*}{Method}& AC& PR& RE&JA  \\
            \hline
            PhaseNet \cite{phasenet}&$78.8$ & $71.3$ & $76.6$ & $-$ \\
            SV-RCNet \cite{svrcnet}&$85.3$&$80.7$&$83.5$&$-$\\
            UATD \cite{ding2023less} & $88.6$ & $86.1$ & $88.0$ & $73.7$ \\
            TeCNO \cite{tecno}& $88.6$&$86.5$&$87.6$&$75.1$\\
            MTRCNet-CL \cite{jin2020multi} & $89.2$ & $86.9$ & $88.0$ & $-$ \\
            Trans-SVNet \cite{trans-svnet}&$90.3$&$90.7$&$88.8$&$79.3$\\
            STAR-Net \cite{chen2023temporal} & $91.2$ & $91.6$ & $89.2$ & $79.5$\\
            OperA \cite{opera}&$91.3$&$-$&$-$&$-$\\
            LoViT \cite{lovit}&$91.5$&$83.1$&$86.5$&$74.2$\\
            SKiT \cite{skit}&$92.5$&$90.9$&$\underline{91.8}$&$\underline{82.6}$\\
                        \rowcolor[rgb]{ .949,  .949,  .949}
SurgPLAN++ Online &\underline{$92.7$}& \underline{$91.1$}&$89.8$&$81.4$\\
                        \rowcolor[rgb]{ .949,  .949,  .949}
SurgPLAN++ Offline&$\boldsymbol{94.1}$&$\boldsymbol{93.3}$&$\boldsymbol{92.9}$&$\boldsymbol{83.5}$\\ \hline
            

            \toprule[1pt]
        \end{tabular}
        \label{cholect50}
    \end{center}
\end{table}


\begin{figure*}[th]
  \centering
  \centerline{\includegraphics[width=0.98\textwidth]{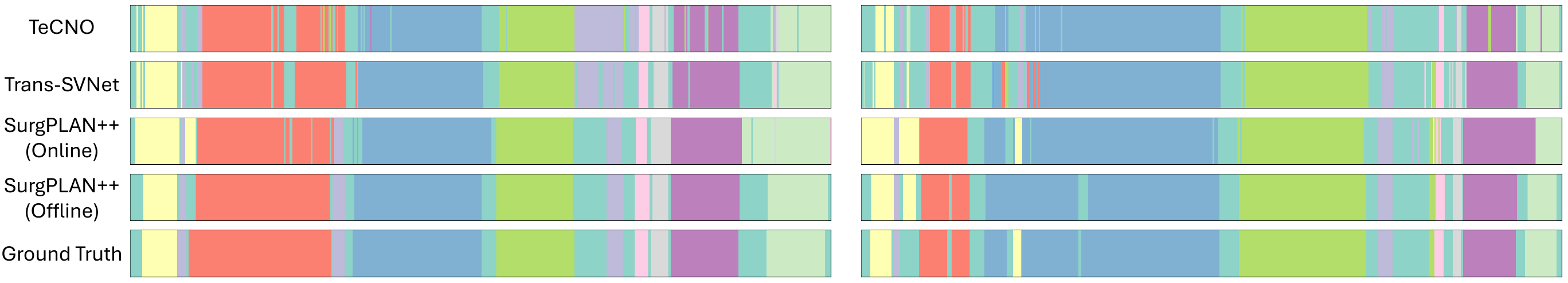}}
\caption{Color-coded ribbon of TeCNO, Trans-SVNet, our SurgPLAN++ and ground truth on the Cataract dataset.}
\label{colorbar}
\end{figure*}

\subsection{Comparison with State-of-the-art Methods}
We compare our SurgPLAN++ with other surgical phase detection models in both Cholec80 and Cataracts datasets.

\noindent \textbf{Comparison on Cholec80 Dataset}. 
To evaluate the performance of surgical phase recognition, we compare SurgPLAN++ offline and online methods with state-of-the-art methods \cite{phasenet,svrcnet,tecno,trans-svnet,chen2023temporal,opera,lovit,skit} on the Cholec80 benchmark. As shown in Table \ref{cholect50}, our SurgPLAN++ with the offline mode reaches the best accuracy and Jaccard score of 94.1\% and 83.5\% among state-of-the-art methods. In particular, our SurgPLAN++ outperforms the SKiT \cite{skit} with a 2.4\% and 1.1\% increase in precision and recall, respectively. This overwhelming performance proves the advantages of the phase localization strategy. 

Furthermore, compared with the online approaches \cite{chen2023temporal,opera,lovit,skit}, the SurgPLAN++ with the online mode also reveals superior accuracy and precision of 92.7\% and 91.1\%. These comparisons further validate the effectiveness of our SurgPLAN++ framework, especially the tailored data augmentation for the online mode of SurgPLAN++.

\noindent \textbf{Comparison on Cataracts Dataset}. 
We further validate the SurgPLAN++ with the open-sourced surgical phase recognition methods \cite{phasenet,svrcnet,tecno,trans-svnet} on the Cataracts dataset. As shown in Table \ref{cataract}, the SurgPLAN++ also reveals a consistent advantage in both online and offline analysis for surgical videos. In particular, our SurgPLAN++ with the offline mode reaches the best accuracy and Jaccard score of 84.3\% and 68.4\%. For the online comparison, our online SurgPLAN++ also has superior accuracy and Jaccard score of 76.7\% and 66.8\%, outperforming the second-best method Trans-SVNet \cite{trans-svnet} with 1.1\% in precision and 5.0\% in recall. These comparisons confirm the advantages of the SurgPLAN++ with both online and offline modes in phase recognition for different types of surgical videos.

\subsection{Qualitative Analysis} 
We further qualitatively compare SurgPLAN++ with the superior approaches in Table \ref{cataract}, \textit{i.e.}, TeCNO \cite{tecno} and Trans-SVNet \cite{trans-svnet} by the color-coded ribbon results on the Cataract dataset. As illustrated in Fig. \ref{colorbar}, the SurgPLAN++ with the offline mode reveals the best performance and is closest to the ground truth. Moreover, the SurgPLAN++ with the online mode also outperforms TeCNO \cite{tecno} and Trans-SVNet \cite{trans-svnet}, especially alleviating the problem of inconsistent predictions of successive frames. Therefore, these qualitative results further confirm the superiority of our SurgPLAN++ in both online and offline surgical phase recognition, by revealing more accurate and continuous prediction intervals.

\begin{table}[t]
    \begin{center}
            \caption{Comparison with with state-of-the-arts on Cataract }
        \begin{tabular}{l| p{1cm}<{\centering} p{1cm}<{\centering} p{1cm}<{\centering} p{1cm}<{\centering} } \hline 
        \toprule[1pt]
            \multirow{1}{*}{Method}& AC& PR& RE&JA  \\
            \hline
            PhaseNet \cite{phasenet} & $68.3$ & $55.4$ & $47.7$& $36.1$  \\
            SV-RCNet \cite{svrcnet}&$70.6$&$57.6$&$50.8$&$38.2$\\
            TeCNO \cite{tecno}& $73.5$&$59.3$&$54.3$&$41.9$ \\
            Trans-SVNet \cite{trans-svnet}&$75.9$&$72.3$&$70.9$&$59.7$\\
                        \rowcolor[rgb]{ .949,  .949,  .949}
SurgPLAN++ Online & $\underline{76.7}$&$\underline{73.4}$&$\underline{75.9}$&$\underline{66.8}$\\
                        \rowcolor[rgb]{ .949,  .949,  .949}
SurgPLAN++ Offline & $\boldsymbol{84.3}$&$\boldsymbol{76.4}$&$\boldsymbol{76.1}$&$\boldsymbol{68.4}$\\ \hline

            \toprule[1pt]
        \end{tabular}
        \label{cataract}
    \end{center}

\end{table}

\subsection{Ablation Study}
To investigate the impact of data augmentation techniques, we conduct a detailed ablation analysis of SurgPLAN++ in the online mode on the Cataract dataset, as shown in Table \ref{tab2}. Note that C-D refers to the Center-Duplication and D-sampling refers to down-sampling. Compared without any data augmentation, by only using the phase of the region proposal that contains the last frame as our prediction, our SurgPLAN++ improves by a large margin, $e.g.$, 35.6\% and 39.9\% in accuracy and Jaccard score, respectively. Compared with only using mirroring or center-duplication, our SurgPLAN++ improves by 28.3\% and 10.6\% in accuracy, and 34.3\% and 8.6\% in Jaccard score, respectively. The down-sampling method can also improve the accuracy and Jaccard score by 2.1\% and 3.8\%. This suggests that by employing data augmentation techniques, SurgPLAN++ transforms the video into a pseudo-complete sequence, thereby enhancing the ability of the phase localization network to capture a segmentation more effectively.

\begin{table}[t]
    \begin{center}
    \caption{Ablation study of SurgPLAN++ for online analysis on Cataract dataset.}
        \begin{tabular}{c c c | c c c c }
            \hline
            \toprule[1pt]
            Mirroring & C-D & D-sampling& AC &  PR &RE&JA \\ 
            \hline
             &&&$41.1$&$43.7$&$39.2$&$26.9$\\ 
             \checkmark&&&$48.4$&$74.4$&$49.5$&$32.5$\\ 
            &\checkmark&&$66.1$&$\textbf{75.8}$&$66.8$&$58.2$ \\
            \checkmark&\checkmark&&$74.6$&$73.2$&$72.2$&$63.0$ \\\checkmark&\checkmark&\checkmark&$\boldsymbol{76.7 }$&$73.4$&$\textbf{75.9}$& $\textbf{66.8}$\\ 
            \toprule[1pt]
        \end{tabular}
        \label{tab2}
    \end{center}
\end{table}

\section{Conclusion}
In this work, we propose a universal SurgPLAN++ framework for both online and offline surgical phase recognition. Different from existing studies that focus on merely online inference and analyze surgical videos as frame-wise classification, our SurgPLAN++ is developed with the principle of temporal detection and predicts phase segments across the entire video through phase proposals. In particular, for online analysis, SurgPLAN++ incorporates a data augmentation strategy to extend the streaming video into a pseudo-complete video. For offline analysis, SurgPLAN++ continuously refines preceding predictions during each online inference step, thereby significantly improving the accuracy of phase recognition. Extensive experiments confirm the superiority of our SurgPLAN++ in both online and offline analysis for surgical videos.
\vspace{6mm}






\bibliographystyle{IEEEtran}
\bibliography{refs}

\begin{thebibliography}{10}
\providecommand{\url}[1]{#1}
\csname url@rmstyle\endcsname
\providecommand{\newblock}{\relax}
\providecommand{\bibinfo}[2]{#2}
\providecommand\BIBentrySTDinterwordspacing{\spaceskip=0pt\relax}
\providecommand\BIBentryALTinterwordstretchfactor{4}
\providecommand\BIBentryALTinterwordspacing{\spaceskip=\fontdimen2\font plus
\BIBentryALTinterwordstretchfactor\fontdimen3\font minus \fontdimen4\font\relax}
\providecommand\BIBforeignlanguage[2]{{%
\expandafter\ifx\csname l@#1\endcsname\relax
\typeout{** WARNING: IEEEtran.bst: No hyphenation pattern has been}%
\typeout{** loaded for the language `#1'. Using the pattern for}%
\typeout{** the default language instead.}%
\else
\language=\csname l@#1\endcsname
\fi
#2}}

\bibitem{chen2021super}
Z.~Chen, X.~Guo, P.~Y. Woo, and Y.~Yuan, ``Super-resolution enhanced medical image diagnosis with sample affinity interaction,'' \emph{IEEE Transactions on Medical Imaging}, vol.~40, no.~5, pp. 1377--1389, 2021.

\bibitem{chen2021diagnose}
Z.~Chen, J.~Zhang, S.~Che, J.~Huang, X.~Han, and Y.~Yuan, ``Diagnose like a pathologist: Weakly-supervised pathologist-tree network for slide-level immunohistochemical scoring,'' in \emph{AAAI}, vol.~35, no.~1, 2021, pp. 47--54.

\bibitem{yang2023hierarchical}
Q.~Yang, Z.~Chen, and Y.~Yuan, ``Hierarchical bias mitigation for semi-supervised medical image classification,'' \emph{IEEE Transactions on Medical Imaging}, vol.~42, no.~8, pp. 2200--2210, 2023.

\bibitem{chen2022instance}
Z.~Chen, J.~Liu, M.~Zhu, P.~Y. Woo, and Y.~Yuan, ``Instance importance-aware graph convolutional network for 3d medical diagnosis,'' \emph{Medical Image Analysis}, vol.~78, p. 102421, 2022.

\bibitem{maier2022surgical}
L.~Maier-Hein, M.~Eisenmann, D.~Sarikaya, K.~M{\"a}rz, T.~Collins, A.~Malpani, J.~Fallert, H.~Feussner, S.~Giannarou, P.~Mascagni, \emph{et~al.}, ``Surgical data science--from concepts toward clinical translation,'' \emph{Medical Image Analysis}, vol.~76, p. 102306, 2022.

\bibitem{xu2024transforming}
H.~Xu, J.~Wu, G.~Cao, Z.~Chen, Z.~Lei, and H.~Liu, ``Transforming surgical interventions with embodied intelligence for ultrasound robotics,'' in \emph{MICCAI}.\hskip 1em plus 0.5em minus 0.4em\relax Springer, 2024, pp. 703--713.

\bibitem{garrow2021machine}
C.~R. Garrow, K.-F. Kowalewski, L.~Li, M.~Wagner, M.~W. Schmidt, S.~Engelhardt, D.~A. Hashimoto, H.~G. Kenngott, S.~Bodenstedt, S.~Speidel, \emph{et~al.}, ``Machine learning for surgical phase recognition: a systematic review,'' \emph{Annals of surgery}, vol. 273, no.~4, pp. 684--693, 2021.

\bibitem{zhai2024artificial}
Y.~Zhai, Z.~Chen, Z.~Zheng, X.~Wang, X.~Yan, X.~Liu, J.~Yin, J.~Wang, and J.~Zhang, ``Artificial intelligence for automatic surgical phase recognition of laparoscopic gastrectomy in gastric cancer,'' \emph{IJCARS}, vol.~19, no.~2, pp. 345--353, 2024.

\bibitem{chen2024asi}
Z.~Chen, Z.~Zhang, W.~Guo, X.~Luo, L.~Bai, J.~Wu, H.~Ren, and H.~Liu, ``Asi-seg: Audio-driven surgical instrument segmentation with surgeon intention understanding,'' in \emph{IROS}.\hskip 1em plus 0.5em minus 0.4em\relax IEEE, 2024, pp. 13\,773--13\,779.

\bibitem{panesar2020promises}
S.~S. Panesar, M.~Kliot, R.~Parrish, J.~Fernandez-Miranda, Y.~Cagle, and G.~W. Britz, ``Promises and perils of artificial intelligence in neurosurgery,'' \emph{Neurosurgery}, vol.~87, no.~1, pp. 33--44, 2020.

\bibitem{wu2024surgbox}
J.~Wu, X.~Liang, X.~Bai, and Z.~Chen, ``Surgbox: Agent-driven operating room sandbox with surgery copilot,'' in \emph{IEEE Big Data}.\hskip 1em plus 0.5em minus 0.4em\relax IEEE, 2024, pp. 2041--2048.

\bibitem{chen2024surgfc}
Z.~Chen, X.~Luo, J.~Wu, D.~T. Chan, Z.~Lei, S.~Ourselin, and H.~Liu, ``Surgfc: Multimodal surgical function calling framework on the demand of surgeons,'' in \emph{BIBM}.\hskip 1em plus 0.5em minus 0.4em\relax IEEE, 2024, pp. 3076--3081.

\bibitem{kirubarajan2022artificial}
A.~Kirubarajan, D.~Young, S.~Khan, N.~Crasto, M.~Sobel, and D.~Sussman, ``Artificial intelligence and surgical education: a systematic scoping review of interventions,'' \emph{Journal of Surgical Education}, vol.~79, no.~2, pp. 500--515, 2022.

\bibitem{tecno}
T.~Czempiel, M.~Paschali, M.~Keicher, W.~Simson, H.~Feussner, S.~T. Kim, and N.~Navab, ``Tecno: Surgical phase recognition with multi-stage temporal convolutional networks,'' in \emph{MICCAI}, 2020, pp. 343--352.

\bibitem{ms-tcn}
G.~J. Farha, Yazan~Abu, ``Ms-tcn: Multi-stage temporal convolutional network for action segmentation,'' in \emph{CVPR}, 2019, pp. 3575--3584.

\bibitem{deepphase}
O.~Zisimopoulos, E.~Flouty, I.~Luengo, P.~Giataganas, J.~Nehme, A.~Chow, and D.~Stoyanov, ``Deepphase: surgical phase recognition in cataracts videos,'' in \emph{MICCAI}, 2018, pp. 265--272.

\bibitem{trans-svnet}
X.~Gao, Y.~Jin, Y.~Long, Q.~Dou, and P.-A. Heng, ``Trans-svnet: Accurate phase recognition from surgical videos via hybrid embedding aggregation transformer,'' in \emph{MICCAI}, 2021, pp. 593--603.

\bibitem{yang2022semi}
Q.~Yang, X.~Liu, Z.~Chen, B.~Ibragimov, and Y.~Yuan, ``Semi-supervised medical image classification with temporal knowledge-aware regularization,'' in \emph{MICCAI}.\hskip 1em plus 0.5em minus 0.4em\relax Springer, 2022, pp. 119--129.

\bibitem{chen2023surgical}
Z.~Chen, Q.~Guo, L.~K. Yeung, D.~T. Chan, Z.~Lei, H.~Liu, and J.~Wang, ``Surgical video captioning with mutual-modal concept alignment,'' in \emph{MICCAI}.\hskip 1em plus 0.5em minus 0.4em\relax Springer, 2023, pp. 24--34.

\bibitem{C3D}
D.~Tran, L.~Bourdev, R.~Fergus, L.~Torresani, and M.~Paluri, ``Learning spatiotemporal features with 3d convolutional networks,'' in \emph{ICCV}, 2015, pp. 4489--4497.

\bibitem{P3D}
Z.~Qiu, T.~Yao, and T.~Mei, ``Learning spatio-temporal representation with pseudo-3d residual networks,'' in \emph{ICCV}, 2017, pp. 5533--5541.

\bibitem{I3D}
X.~Weng and K.~Kitani, ``Learning spatio-temporal features with two-stream deep 3d cnns for lipreading,'' \emph{arXiv preprint arXiv:1905.02540}, 2019.

\bibitem{slowfast}
C.~Feichtenhofer, H.~Fan, J.~Malik, and K.~He, ``Slowfast networks for video recognition,'' in \emph{ICCV}, 2019, pp. 6202--6211.

\bibitem{phasenet}
A.~P. Twinanda, D.~Mutter, J.~Marescaux, M.~de~Mathelin, and N.~Padoy, ``Single-and multi-task architectures for surgical workflow challenge at m2cai 2016,'' \emph{arXiv preprint arXiv:1610.08844}, 2016.

\bibitem{MSTCN}
S.~R. Sekaran, Y.~H. Pang, G.~F. Ling, and O.~S. Yin, ``Mstcn: A multiscale temporal convolutional network for user independent human activity recognition,'' \emph{F1000Research}, vol.~10, 2021.

\bibitem{TMR}
Y.~Jin, Y.~Long, C.~Chen, Z.~Zhao, Q.~Dou, and P.-A. Heng, ``Temporal memory relation network for workflow recognition from surgical video,'' \emph{IEEE Transactions on Medical Imaging}, vol.~40, no.~7, pp. 1911--1923, 2021.

\bibitem{svrcnet}
Y.~Jin, Q.~Dou, H.~Chen, L.~Yu, J.~Qin, C.-W. Fu, and P.-A. Heng, ``Sv-rcnet: workflow recognition from surgical videos using recurrent convolutional network,'' \emph{IEEE Transactions on Medical Imaging}, vol.~37, no.~5, pp. 1114--1126, 2017.

\bibitem{actionformer}
C.-L. Zhang, J.~Wu, and Y.~Li, ``Actionformer: Localizing moments of actions with transformers,'' in \emph{ECCV}, 2022, pp. 492--510.

\bibitem{tridet}
D.~Shi, Y.~Zhong, Q.~Cao, L.~Ma, J.~Li, and D.~Tao, ``Tridet: Temporal action detection with relative boundary modeling,'' in \emph{CVPR}, June 2023, pp. 18\,857--18\,866.

\bibitem{softnms}
N.~Bodla, B.~Singh, R.~Chellappa, and L.~S. Davis, ``Soft-nms--improving object detection with one line of code,'' in \emph{ICCV}, 2017, pp. 5561--5569.

\bibitem{twinanda2016endonet}
A.~P. Twinanda, S.~Shehata, D.~Mutter, J.~Marescaux, M.~De~Mathelin, and N.~Padoy, ``Endonet: a deep architecture for recognition tasks on laparoscopic videos,'' \emph{IEEE Transactions on Medical Imaging}, vol.~36, no.~1, pp. 86--97, 2016.

\bibitem{dataset}
\BIBentryALTinterwordspacing
H.~Al~Hajj, M.~Lamard, P.-h. Conze, B.~Cochener, and G.~Quellec, ``Cataracts,'' 2021. [Online]. Available: \url{https://dx.doi.org/10.21227/ac97-8m18}
\BIBentrySTDinterwordspacing

\bibitem{al2019cataracts}
H.~Al~Hajj, M.~Lamard, P.-H. Conze, S.~Roychowdhury, X.~Hu, G.~Mar{\v{s}}alkait{\.e}, O.~Zisimopoulos, M.~A. Dedmari, F.~Zhao, J.~Prellberg, \emph{et~al.}, ``Cataracts: Challenge on automatic tool annotation for cataract surgery,'' \emph{Medical Image Analysis}, vol.~52, pp. 24--41, 2019.

\bibitem{luo2024surgplan}
X.~Luo, Y.~Pang, Z.~Chen, J.~Wu, Z.~Zhang, Z.~Lei, and H.~Liu, ``Surgplan: Surgical phase localization network for phase recognition,'' in \emph{ISBI}.\hskip 1em plus 0.5em minus 0.4em\relax IEEE, 2024.

\bibitem{ding2023less}
X.~Ding, X.~Yan, Z.~Wang, W.~Zhao, J.~Zhuang, X.~Xu, and X.~Li, ``Less is more: Surgical phase recognition from timestamp supervision,'' \emph{IEEE Transactions on Medical Imaging}, 2023.

\bibitem{jin2020multi}
Y.~Jin, H.~Li, Q.~Dou, H.~Chen, J.~Qin, C.-W. Fu, and P.-A. Heng, ``Multi-task recurrent convolutional network with correlation loss for surgical video analysis,'' \emph{Medical Image Analysis}, vol.~59, p. 101572, 2020.

\bibitem{chen2023temporal}
Z.~Chen, Y.~Zhai, J.~Zhang, and J.~Wang, ``Surgical temporal action-aware network with sequence regularization for phase recognition,'' in \emph{BIBM}.\hskip 1em plus 0.5em minus 0.4em\relax IEEE, 2023, pp. 1836--1841.

\bibitem{opera}
T.~Czempiel, M.~Paschali, D.~Ostler, S.~T. Kim, B.~Busam, and N.~Navab, ``Opera: Attention-regularized transformers for surgical phase recognition,'' in \emph{MICCAI}.\hskip 1em plus 0.5em minus 0.4em\relax Springer, 2021, pp. 604--614.

\bibitem{lovit}
Y.~Liu, M.~Boels, L.~C. Garcia-Peraza-Herrera, T.~Vercauteren, P.~Dasgupta, A.~Granados, and S.~Ourselin, ``Lovit: Long video transformer for surgical phase recognition,'' \emph{arXiv preprint arXiv:2305.08989}, 2023.

\bibitem{skit}
Y.~Liu, J.~Huo, J.~Peng, R.~Sparks, P.~Dasgupta, A.~Granados, and S.~Ourselin, ``Skit: a fast key information video transformer for online surgical phase recognition,'' in \emph{ICCV}, 2023, pp. 21\,074--21\,084.

\end{thebibliography}

\end{document}